\documentclass[nohyperref]{article}

\usepackage{microtype}
\usepackage{graphicx}
\usepackage{subfigure}
\usepackage{booktabs} 
\usepackage{hyperref}

\usepackage[accepted]{icml2023}

\usepackage{amsmath}
\usepackage{amssymb}
\usepackage{mathtools}
\usepackage{amsthm}
\usepackage{multirow}
\usepackage{braket}
\usepackage{wrapfig}

\usepackage[capitalize,noabbrev]{cleveref}

\theoremstyle{plain}

\theoremstyle{definition}

\theoremstyle{remark}

\usepackage[textsize=tiny]{todonotes}
\usepackage{color,soul}

\icmltitlerunning{Efficient and Equivariant Graph Networks for Predicting Quantum Hamiltonian}

\begin{document}

\twocolumn[
\icmltitle{Efficient and Equivariant Graph Networks for Predicting Quantum Hamiltonian}
\icmlsetsymbol{equal}{*}


\begin{icmlauthorlist}
\icmlauthor{Haiyang Yu}{tamucs}
\icmlauthor{Zhao Xu}{tamucs}
\icmlauthor{Xiaofeng Qian}{tamumaterial,tamuece,tamuphys,equal}
\icmlauthor{Xiaoning Qian}{tamucs,tamuece,equal}
\icmlauthor{Shuiwang Ji}{tamucs,equal}
\end{icmlauthorlist}

\icmlaffiliation{tamucs}{Department of Computer Science \& Engineering, Texas A\&M University, TX, USA}
\icmlaffiliation{tamuece}{Department of Electrical \& Computer Engineering, Texas A\&M University, TX, USA}
\icmlaffiliation{tamumaterial}{Department of Materials Science \& Engineering, Texas A\&M University, TX, USA}
\icmlaffiliation{tamuphys}{Department of Physics \& Astronomy, Texas A\&M University, TX, USA}

\icmlcorrespondingauthor{Shuiwang Ji}{sji@tamu.edu}

\icmlkeywords{Machine Learning, ICML}
\vskip 0.3in
]
\newcommand{\red}[1]{\textcolor{red}{#1}}



\printAffiliationsAndNotice{\icmlEqualSeniorContribution} 

\begin{abstract}

We consider the prediction of the Hamiltonian matrix, which finds use in quantum chemistry and condensed matter physics. Efficiency and equivariance are two important, but conflicting factors. In this work, we propose a SE(3)-equivariant network, named QHNet, that achieves efficiency and equivariance. Our key advance lies at the innovative design of QHNet architecture, which not only obeys the underlying symmetries, but also enables the reduction of number of tensor products by 92\%. In addition, QHNet prevents the exponential growth of
channel dimension when more atom types are involved.
We perform experiments on MD17 datasets, including four molecular systems. Experimental results show that our QHNet can achieve comparable performance to the state of the art methods at a significantly faster speed. Besides, our QHNet consumes 50\% less memory due to its streamlined architecture. Our code is publicly available as part of the AIRS library (\url{https://github.com/divelab/AIRS}).
\end{abstract}



\section{Introduction}

Deep learning has achieved significant progress in computational quantum chemistry in recent years. Existing deep learning methods have demonstrated their efficiency and expressiveness in tackling various challenging quantum mechanical simulation tasks. For example, deep graph learning methods can now accurately predict quantum properties of a molecule, such as molecular energy and the HOMO-LUMO gap~\cite{schutt2017schnet, liu2021spherical, gasteiger_dimenet_2020, klicpera2020fast, wang2022advanced, liu2021fast, wang2022comenet, brandstettergeometric, yan2022periodic}. Recent deep generative models have also shown to be capable of generating new materials and molecules by faithfully learning the distribution of their structures~\cite{simm2020generative, mansimov2019molecular, xu2021an, shi2021learning, xu2021learning, ganea2021geomol, luo2021autoregressive}. 
Recent efforts on modeling interaction of two and more molecules also shed light on protein-ligand docking for drug development~\cite{corso2022diffdock, ganea2021independent, stark2022equibind, zhang2022e3bind, lu2022tankbind, jiang2022predicting, liu2022generating}.
Inspired by all these advancements,  
we aim to predict a more fundamental target in computational physics, quantum tensors, by developing a new deep graph learning model framework in this work. 

Quantum tensors, such as Hamiltonian matrix, eigen wavefunctions, and eigen energies, can be used to describe molecular systems and their quantum states. Since quantum tensors contain the most critical information about molecular systems, many molecular properties can be directly derived from quantum tensors and wavefunctions. Unfortunately, obtaining precise quantum tensors is at considerably high cost. Density  functional theory (DFT) \cite{Hohenberg64,dft} and \textit{ab initio} quantum chemistry methods \cite{szabo2012modern} are routinely used to calculate  electronic wavefunctions, charge density, and total energy of molecules and solids. However, first-principles methods are computationally very expensive, limiting their use in small systems. Therefore, deep learning is believed to have the potential to accelerate quantum mechanical simulations if it can accurately and reliably predict quantum tensors \cite{schutt2019unifying, unke2021se} . 

\begin{figure}[t]
    \vskip 0.2in
    \begin{center}
    \centerline{\includegraphics[width=0.48\textwidth]{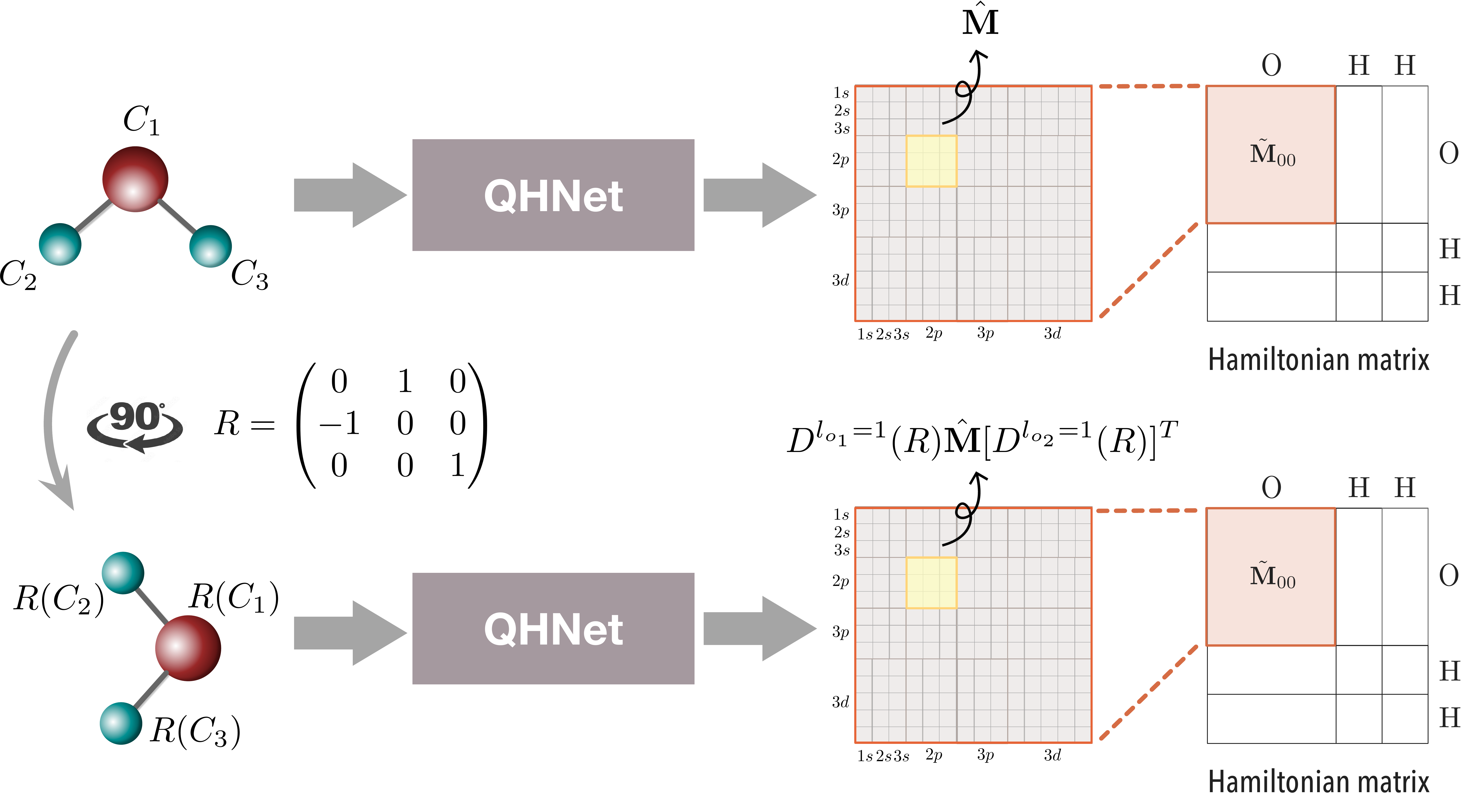}}
    \caption{The rotation equivariance relationship between the input molecule and output Hamiltonian matrix.}
    \label{fig:matrix-equi}
    \end{center}
    \vskip -0.4in
\end{figure}

Unlike invariant molecular properties such as energy and equivariant properties such as atomic forces, quantum tensors possess a higher rotation order to reflect the exact rotation of a molecule. Therefore, developing deep learning methods to predict quantum tensors is challenging and requires elaborate designs of model architectures. SE(3)-equivariant graph neural networks \cite{satorras2021egnn, schutt2021painn, thomas2018tensor, anderson2019cormorant, gasteiger2021gemnet} have the promising potential in predicting quantum tensors since they ensure the equivariance of permutation, translation, and rotation. Output quantum tensors are guaranteed to be permuted, translated, and rotated in the same way as the input molecule shown in Figure \ref{fig:matrix-equi}.  However, the efficiency of SE(3)-equivariant models are generally low as a trade-off for equivariance. In this work, we propose an efficient and equivariant graph network, named QHNet, for predicting quantum tensors including Hamiltonian matrices.

The high efficiency of the proposed QHNet is mainly because QHNet adopts much fewer tensor product (TP) operations than existing SE(3)-equivariant graph networks. Note that TP operation causes most of the time overheads though they are critical for learning quantum tensors. As a result, QHNet boosts efficiency by more than three times and reduces 50\% of GPU memory, which is a significant advance compared to existing SE(3)-equivariant networks. With such an efficient architecture, our QHNet surprisingly achieves comparable performance to the state of the art methods. 

In addition to better efficiency and prediction performance, QHNet has another advantage of more flexible applicability. The expansion module of QHNet outputs intermediate blocks with a fixed shape for all node pairs despite atom types. The fixed shape is predefined by full orbitals needed for a particular dataset. Then, the pairwise blocks of every two atoms are extracted to build the quantum tensor according to specific atomic orbitals. This design enables an easy extension to cases where more atom types are involved, {\it i.e.}, the model architecture is not affected by different types of atoms. Therefore, given the advantages of efficacy, efficiency, and applicability, our QHNet is an excellent choice for predicting quantum tensors such as quantum Hamiltonian, in more general molecular systems.

\section{Background and Related Work}
\subsection{Quantum Tensors in Density Functional Theory}

Quantum mechanical methods such as first-principles density functional theory (DFT) \cite{Hohenberg64,dft} and \textit{ab initio} quantum chemistry methods \cite{szabo2012modern}, are widely used to calculate the electronic structure of solids and molecules, such as electronic wavefunctions, charge density, total energy etc. These quantities can in turn be used to derive many other physical and chemical properties, for examples, electronic, mechanical, optical, magnetic, and catalytic properties of molecules and solids. 
DFT maps a many-body interacting system onto a many one-body non-interacting system, allowing to efficiently solve the Schr\"{o}dinger equation for complex materials and molecules. In principle, DFT with exact exchange-correlation energy functional yields exact ground-state density and total energy. The Schr\"{o}dinger equation for an $n$-electron interacting system is given by, 
\begin{equation}
    \boldsymbol{\hat{H}}\Psi\left(\boldsymbol{r}_1, \ldots, \boldsymbol{r}_{n}\right) = {E} \Psi\left(\boldsymbol{r}_1, \ldots, \boldsymbol{r}_{n}\right),
    \label{eq:sch}
\end{equation}
where $\Psi$ is the total wavefunction of all electrons, $\boldsymbol{\hat{H}}$ is the Hamiltonian operator, and ${E}$ is the total energy. Here, the ionic degree of freedom is not considered. 
DFT takes use of a set of single electronic wavefunctions $\{ \psi_i (\mathbf{r})\}$ to approximate the electronic density for the many-electron system.
In molecular systems, $\psi_i$ usually represents the $i$-th molecular orbital. Then $\psi_i$ is approximated by a linear combination of predefined basis set $\phi_j$ as
\begin{equation}
     \psi_i(\boldsymbol{r}) = \sum_j \mathbf{C}_{ij} \phi_j (\boldsymbol{r}),
     \label{eq:wavefunction}
\end{equation} 
where $\mathbf{C} \in \mathbb{R}^{n \times n}$ denotes the molecular orbital coefficients of interest, although it could have complex values for solids etc. Derived from the Schr\"{o}dinger equation in Eq.~\eqref{eq:sch}, the coefficients $\mathbf{C}$ for the molecular ground state satisfy the eigenvalue decomposition equation \cite{szabo2012modern}, 
\begin{equation}
    \mathbf{H} \mathbf{C} = \boldsymbol{\epsilon} \mathbf{S} \mathbf{C},
    \label{eq:roothaan}
\end{equation} 
where $\mathbf{H}$ is the Hamiltonian matrix with $\mathbf{H}_{i j}=
\braket{\phi_i|\hat{\mathbf{H}}| \phi_j} =\int \phi_i^*(\boldsymbol{r}) \hat{\mathbf{H}} \phi_j(\boldsymbol{r}) d\boldsymbol{r}$,
$\mathbf{S}$ is the overlap matrix with $\boldsymbol{S}_{ij} = \braket{\phi_i|\phi_j}  = \int \phi_i^*(\boldsymbol{r}) \phi_j(\boldsymbol{r}) d\boldsymbol{r}$ which represents the integral of a pair of the predefined basis functions, and each element in the diagonal matrix $\boldsymbol{\epsilon}$ denotes the energy of the corresponding molecular orbital. Note that $\mathbf{H}, \mathbf{S}, \mathbf{\epsilon} \in \mathbb{R}^{n_o \times n_o}$where $n_o$ denotes the number of orbitals used in DFT calculations.

The ultimate goal of DFT is to calculate these quantum tensors $\mathbf{H}$, $\mathbf{C}$, and $\mathbf{\epsilon}$, from which other properties such as total energy and atomic forces etc. can be readily computed. By utilizing the self-consistent field~(SCF) methods~\cite{payne1992rmp, cances2000convergence, kudin2002black}, DFT can approximate the solution iteratively with a time complexity of $O(n^3)$ for each step, resulting in significant cost for running DFT simulations for large systems. This is particularly true when multiple iterations have to be conducted to obtain the final quantum tensors in settings in which a high level of accuracy is required. With the recent advances and successes of machine learning approaches for scientific computing, we here propose to develop deep learning models to predict quantum tensors within a reasonable level of accuracy, thereby accelerating the optimization process in the electronic structure calculations with better accuracy.




\subsection{Related Work}








In recent years, graph neural networks have yielded promising results in quantum chemistry by solving problems such as precise prediction of molecular properties~\cite{gilmer2017mpnn, liu2021spherical, se3transformers, nequip, schutt2021painn, godwin2021noisynode, satorras2021egnn, unke2019physnet, gasteiger_dimenet_2020, qiao2020orbnet}. A molecule usually has various quantum properties that describe the molecule from different perspectives. We divide these properties into three categories based on their rotation orders. A typical example of the first category is molecular energy, including the total energy and the HOMO-LUMO gap, as molecular energy is invariant to molecular rotation. The force falls into the second category because force vectors rotate as a molecule rotates. Thus, force predictions have to be equivariant to rotations of the same order. The third category includes a higher rotation order of the matrix that indicates the exact rotation. The Hamiltonian matrix is an example in this category and is more challenging to predict. To predict molecular properties with different rotation orders, different variants of invariant and equivariant GNNs have been proposed accordingly. 

Invariant GNNs use relative geometric information of molecules as input and predict invariant molecular properties. For instance, SchNet~\cite{schutt2018schnet} considers pairwise distances when performing message passing, while DimeNet~\cite{gasteiger_dimenet_2020} also uses angles in addition to distance. SphereNet~\cite{liu2021spherical} and ComENet \cite{wang2022comenet} incorporate distances, angles, and torsion angles to build more informative representations. 


Equivariant GNNs use equivariant features and specific model architectures to predict the roto-equivariant molecular properties. For example, PaiNN~\cite{schutt2021painn} maintains the equivariant features of rotation order $\ell=1$ in the model architecture. Tenor field networks, SE3-transformers, and NequIP~\cite{thomas2018tensor, se3transformers, nequip} use tensor products to incorporate lifted higher-order of equivariant features while prediction targets could be either invariant or equivariant with order $\ell=1$.


To predict high-order quantum tensors, SchNorb~\cite{schutt2019unifying} follows SchNet~\cite{schutt2018schnet} to consider pairwise distances and incorporates direction information by applying pairwise direction vector on pairwise interaction features. It then constructs blocks of the Hamiltonian matrix. SchNorb considers system coordinates through the pairwise direction vector. However, it does not provide guarantee on yielding an equivariant matrix. 
In PhiSNet~\cite{unke2021se}, tensor products~\cite{thomas2018tensor} are used to ensure equivariance. In this method, the equivariant atomic representation network is applied to extract equivariant features for each atom. The mixing layers are then applied to construct equivariant representations for each node pair. The Hamiltonian matrix is finally constructed with irreducible representations collected from the mapping from orbital interactions to channel indices.
The DeepH~\cite{deeph} uses the 3D GNN to predict the invariant Hamiltonian matrix block. It then applies the Wigner-D matrix to the predicted invariant matrix, thereby ensuring rotation equivariance.

\section{Methodology}
In this section, we describe our proposed SE(3) equivariant graph neural network, QHNet, for quantum tensor prediction tasks.

\subsection{Tensor Field Networks}
The tensor field network (TFN)~\cite{thomas2018tensor} is one of the commonly-used equivariant neural network architectures that achieve 3D rotation, translation, and permutation equivariance. TFN uses tensor product to combine two irreducible representations $u$ and $v$ of the rotation orders 
$\ell_1$ and $\ell_2$ with the Clebsch-Gordan (CG) coefficients~\cite{griffiths2018introduction} to produce a new irreducible representation of order $\ell_3$ as
\begin{equation*}
    (u^{\ell_1} \otimes v^{\ell_2})_{m_3}^{\ell_3} = \sum_{m_1=-\ell_1}^{\ell_1} \sum_{m_2=-\ell_2}^{\ell_2} C^{(\ell_3, m_3)}_{(\ell_1, m_1), (\ell_2, m_2)} u_{m_1}^{\ell_1} v_{m_2}^{\ell_2}, 
\end{equation*} 
where $C$ denotes the CG matrix, $\ell_3$ satisfies $|\ell_1 - \ell_2| \leq \ell_3 \leq \ell_1 + \ell_2$, and $ \ell_1, \ell_2, \ell_3 \in \mathbb{N}$. Note that $m$ denotes the $m$-th element in the irreducible representation with $-\ell \leq m \leq \ell$ and $m \in \mathbb{N}$. In the TFN~\cite{thomas2018tensor}, each layer is composed of filter, convolution, self-interaction and nonlinear activation modules. First, spherical harmonic filters $Y$ are applied to the node pair direction $\hat{r}_{ij}$, then combined with the pairwise distance $r_{ij}$ to obtain the filter outputs $F$ as
\begin{equation}
    F_{cm}^{(\ell\textsubscript{in}, \ell_f)}(r_{ij}, \hat{r}_{ij}) = R_c^{(\ell\textsubscript{in}, \ell_f)}(r_{ij}) Y^{\ell_f}_m(\hat{r}_{ij}),
    \label{eq:tfn_filter}
\end{equation}
where $R$ is a multiple layer perceptron~(MLP) that takes the embedding of pairwise distance as input, and $c$ is the channel index.
Then the convolution collects information from the irreducible representations of other nodes and the pairwise spherical harmonics through the tensor product as 
\begin{equation}
     \tilde{V}_{i}^{\ell\textsubscript{out}} = \sum_j (F^{(\ell\textsubscript{in}, \ell_f)}(r_{ij}, \hat{r}_{ij}) \otimes V_{j}^{\ell\textsubscript{in}})^{\ell\textsubscript{out}}
\end{equation}
with $|\ell\textsubscript{in} - \ell_{f}| \leq \ell\textsubscript{out} \leq \ell\textsubscript{in} + \ell_{f}$, and $\ell\textsubscript{in}, \ell\textsubscript{out}, \ell_{f} \in \mathbb{N}$. 
Then the self-interaction is a linear layer that combines features across the channel dimension to obtain irreducible representations with the same order $\ell$ as
\begin{equation}
    V^{\ell}_{icm} = f\textsubscript{linear}(V_i^{\ell})_{cm} = \sum_c w_{cc'} V^{\ell}_{ic'm}.
\end{equation}
In the final nonlinear layer, a nonlinear function is applied to the invariant irreducible representations of $\ell=0$. Meanwhile, the average norm $\sum_c \| V_{ic}^{\ell} \|$ is calculated over the channel for irreducible representations with $\ell>1$, and $V_{i}^{\ell}$ is normalized with this average norm.

\begin{figure*}[ht]
    \vskip 0.2in
    \centering
    \includegraphics[width=0.95\textwidth]{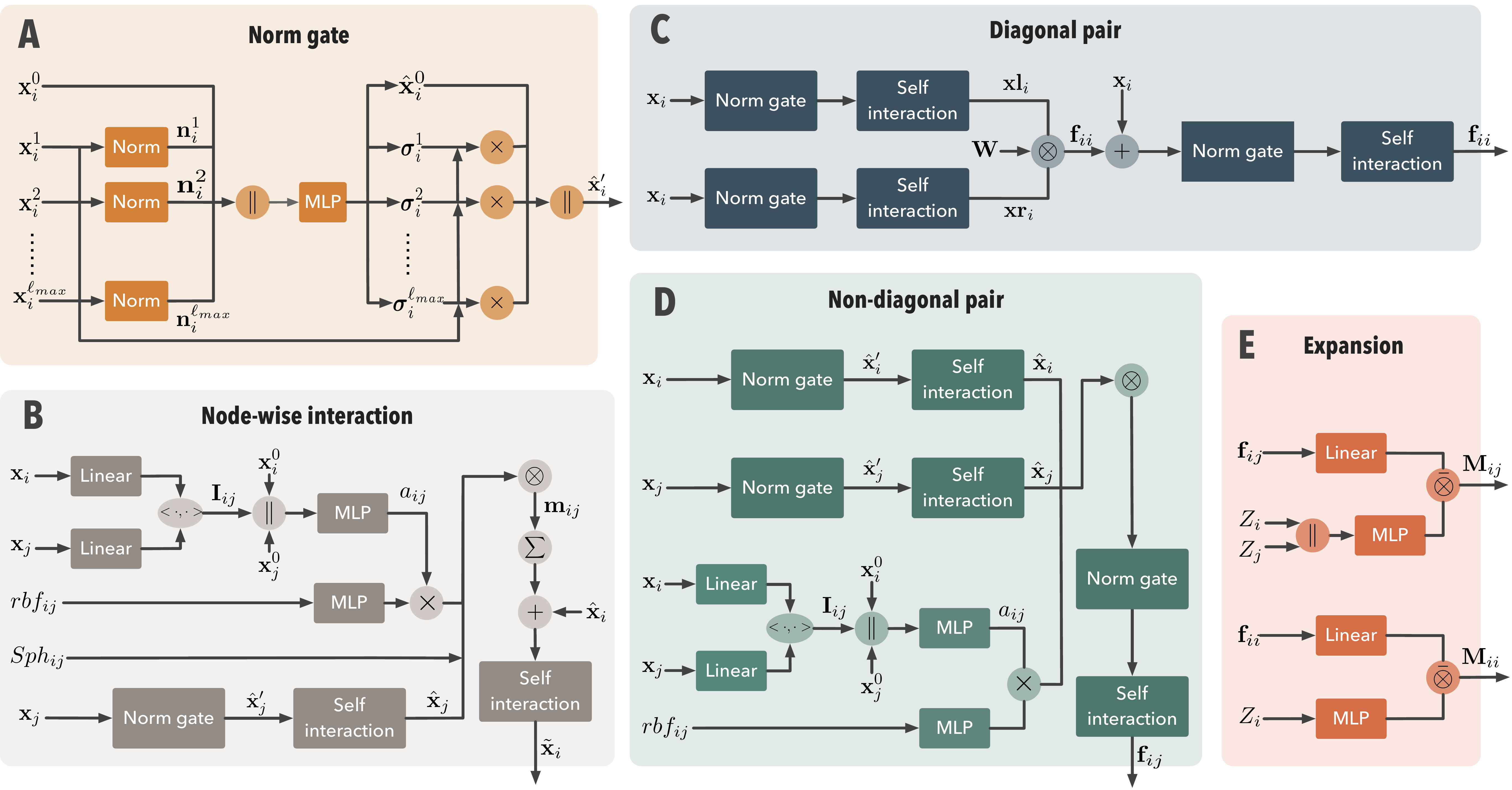}
    \caption{The message passing schema of our model. \textbf{A:} Norm gate. The norm of node representations $\mathbf{x}$ at different orders are concatenated and fed to an MLP to calculate scale factor $\sigma$ at each order. Irreducible representations at different orders are rescaled with these factors and then be concatenated to output $\hat{\mathbf{x}}'$. Note that rescaling only applies to orders higher than 0. \textbf{B:} Node-wise interaction layer. The inner product of linearly transformed representations of nodes $i$ and $j$ is concatenated with their irreducible representation at order $\ell=0$ and then fed to an MLP to calculate the attentive score $a_{ij}$. Next, the spherical harmonics of pairwise direction $\hat{r}_{ij}$, radius basis function (RBF) transformed pairwise distance $r_{ij}$, and attentive score $a_{ij}$ are multiplied to produce the filter. A norm gate followed by a self-interaction layer is applied to $\mathbf{x}_j$ to produce $\hat{\mathbf{x}}_j$ which is used in a tensor product with the filter to obtain message $\mathbf{m}_{ij}$. Finally, messages are aggregated with $\hat{\mathbf{x}}_i$ and updated to output $\tilde{\mathbf{x}}_i$. \textbf{C:} Diagonal pair. Two norm gates with self-interaction are applied to node irreducible representation $\mathbf{x}_i$ in parallel to obtain $\mathbf{xl}_i$ and $\mathbf{xr}_i$. Then, a tensor product with parameters $\mathbf{W}$ is used to produce the diagonal pair representation $\mathbf{f}_{ii}$. The $\mathbf{f}_{ii}$ is skip-connected with $\mathbf{x}_i$ and processed by another norm gate with self-interaction before being outputted. \textbf{D:} Non-diagonal pair. The attentive score $a_{ij}$ of nodes $i$ and $j$ is multiplied with the RBF and MLP transformed pairwise distance $r_{ij}$ to produce the filter. Two norm gates with self-interaction are applied to $\mathbf{x}_i$ and $\mathbf{x}_j$ separately to obtain $\hat{\mathbf{x}}_i$ and $\hat{\mathbf{x}}_j$. Then, the non-diagonal pair representation $\mathbf{f}_{ij}$ is produced by a tensor product applying to $\hat{\mathbf{x}}_i$, $\hat{\mathbf{x}}_j$, and $a_{ij}$. The $\mathbf{f}_{ij}$ is outputted after going through another norm gate with self-interaction. \textbf{E:} Expansion module. For a non-diagonal pair, node embeddings of atoms $i$ and $j$ are concatenated and transformed by an MLP, and then fed into an inverse of tensor product with linearly transformed pair representation $\mathbf{f}_{ij}$ to output the intermediate block matrix $\mathbf{M}_{ij}$ in a fixed shape. For a diagonal pair, inputs of expansion are pair representation $\mathbf{f}_{ii}$ and embedding of node $i$. Output is the intermediate block $\mathbf{M}_{ii}$.}
    \vskip -0.2in
    \label{fig:message-passing}
\end{figure*}

\subsection{Norm Gate}
The irreducible representations can be represented as the multiplication of their norm and direction vectors. When the norm of $\mathbf{x}_i$ is larger than other nodes, the message from node $i$ has greater impact on node messaging updating. Instead of the layer norm operation in the original TFN layer, we perform norm gate learning to rescale the norm of irreducible representations before applying message passing and tensor product. The norm of order $\ell$ representations for node $i$, $\mathbf{x}_i^{\ell}$, is defined as 
\begin{equation}
    \mathbf{n}_i^{\ell} = \| \mathbf{x}_i^{\ell} \|. 
\end{equation}
We apply an MLP to learn the norm of new irreducible representations along with the new irreducible representations with order 0 as
\begin{equation}
    \hat{\mathbf{x}}_i^0, \boldsymbol{\sigma}_{i}^{1}, \dots \boldsymbol{\sigma}_{i}^{\ell\textsubscript{max}} = f\textsubscript{MLP}(\mathbf{x}_i^{0}, \mathbf{n}_{i}^{1}, \dots, \mathbf{n}_{i}^{\ell\textsubscript{max}}). 
\end{equation}
The irreducible representations with order higher than 0 are rescaled with the factor
\begin{equation}
    \hat{\mathbf{x}}_{im}^{' \ell} = \boldsymbol{\sigma}_{i}^{\ell} \mathbf{x}_{im}^{\ell}.
\end{equation}
Since $\mathbf{x}^0$ and the norm $\mathbf{n}_i^{\ell}$ are SE(3) invariant, the scale $\boldsymbol{\sigma}_i$ is also invariant. Therefore, the rescaled irreducible representations $\hat{\mathbf{x}}^{\ell}_i$ have the same SE(3) equivariance as $\mathbf{x}^{\ell}_i$.
The illustration of norm gate is shown in Figure~\ref{fig:message-passing}.A.

Usually, a self-interaction layer is applied after the norm gate to produce the representation $\hat{\mathbf{x}}_{i}^{\ell}$ as
\begin{equation}
    \hat{\mathbf{x}}_{i}^{\ell} = f\textsubscript{linear}(\hat{\mathbf{x}}_{i}^{' \ell}).
\end{equation}
\subsection{Node-Wise Interaction Layer}

The filter module in the TFN layer controls influence of messages from other nodes. It takes pairwise distances as input and induces most of parameters in the TFN layer. Intuitively, when nodes are far from each other, their influence on each other should be weak. 

In our node-wise interaction layers as illustrated in Figure~\ref{fig:message-passing}.B, we further consider the similarity of pairwise nodes. We calculate the inner product of the pairwise irreducible representations of order $\ell$ as 
\begin{equation}
        \mathbf{I}_{ij}^{\ell} = \braket{f\textsubscript{linear}(\mathbf{x}_i^{\ell}), f\textsubscript{linear}(\mathbf{x}_j^{\ell})}.
\end{equation} 
To increase expressive power, a self-interaction linear layer is applied before the inner product. The pairwise cosine similarity of irreducible representations are then fed into a MLP along with the irreducible representations of order $0$ to calculate attentive scores as
\begin{equation}
    a_{ij} = f_{\textsubscript{MLP}}(\mathbf{x}_i^{0}, \mathbf{x}_j^{0}, \mathbf{I}_{ij}^{1}, \dots, \mathbf{I}_{ij}^{\ell\textsubscript{max}}).
    \label{eq:attentive-scores}
\end{equation}
Since $\mathbf{x}^0$ and the inner product $I_{ij}$ are invariant, the attentive scores are also SE(3) invariant.

Even though the filter in the TFN layer assigns a weight for each $(\ell\textsubscript{in}, \ell_f)$ according to Eq.~\eqref{eq:tfn_filter}, the weights are the same when the order of output features $\ell\textsubscript{out}$ are different. To explicitly express the difference, we extend the filter to assign weights for each path $(\ell\textsubscript{in}, \ell_f, \ell\textsubscript{out})$. Furthermore, the filter assigns an attentive score for each path to consider pairwise similarities. Formally,  
\begin{align}
     F_{cm}^{(\ell\textsubscript{in}, \ell_f, \ell\textsubscript{out})}&(r_{ij}, \hat{r}_{ij}) = \nonumber \\
    & a_{ijc}^{(\ell\textsubscript{in}, \ell_f, \ell\textsubscript{out})} 
    R_c^{(\ell\textsubscript{in}, \ell_f, \ell\textsubscript{out})}(r_{ij}) Y^{\ell_f}_m(\hat{r}_{ij}).
    \label{eq:filter}
\end{align}
Note that $c$ denotes the channel index and $m$ indexes the irreducible representations.

In the next step, we calculate the message $\mathbf{m}_{ij}$ from node $j$ to node $i$. The irreducible representations $\mathbf{x}$ are first rescaled with a norm gate and self-interaction layer to obtain $\hat{\mathbf{x}}$. Then the rescaled representations cooperates with the filter to produce $\mathbf{m}_{ij}$ as
\begin{equation}
      \mathbf{m}_{ijc}^{\ell\textsubscript{out}} = \sum_{\ell_f, \ell\textsubscript{in}}
      (F_c^{(\ell\textsubscript{in}, \ell_f, \ell\textsubscript{out})}(r_{ij}, \hat{r}_{ij}) \otimes (\hat{\mathbf{x}}_{jc}^{\ell\textsubscript{in}}))^{\ell\textsubscript{out}}. 
\end{equation}

The output irreducible representations $\tilde{\mathbf{x}}$ are then obtained by aggregating the message $\mathbf{m}_{ij}$ and the self-connection and further updated with a self-interaction layer as
\begin{equation}
    \tilde{\mathbf{x}}_i^{\ell} = f\textsubscript{linear}(\hat{\mathbf{x}}_i^{\ell} + \sum_{j} \mathbf{m}_{ij}^{\ell}).
\end{equation}

\vspace{-5mm}
\subsection{Building Quantum Tensor}
\label{sec:buil_block}

Quantum tensors like Hamiltonian matrix characterize the pairwise relationship between atoms in molecular systems, and they can be divided into diagonal and non-diagonal blocks. The diagonal blocks consider Hamiltonian operators on a single atom, while the non-diagonal blocks are for pairs of two atoms.
In this subsection, we explain how to produce pairwise irreducible representations and then build target quantum matrix block-by-block using tensor expansion with tensor product filters.

\paragraph{Non-Diagonal Pair.} For each pair of non-diagonal nodes, a tensor product filter controls the influence of their irreducible representations onto node pair irreducible representation $\mathbf{f}_{ij}$. 
As shown in Figure~\ref{fig:message-passing}.D, the filter first calculates attentive scores $a_{ij}$ in the same way as Eq.~\eqref{eq:attentive-scores} to consider pairwise similarity $\mathbf{I}_{ij}^{l}$ along with the irreducible representations of order $0$. Next, it combines the attentive scores with $R(r_{ij})$ to produce a weight for each path $(l_{in_i}, l_{in_j}, l_{out})$ as
\begin{align*}
    F_{cm}^{(\ell\textsubscript{in}_i, \ell\textsubscript{in}_j, \ell\textsubscript{out})}&(r_{ij}) = a_{ijc}^{(\ell\textsubscript{in}_i, \ell\textsubscript{in}_j, \ell\textsubscript{out})} R_{cm}^{(\ell\textsubscript{in}_i, \ell\textsubscript{in}_j, \ell\textsubscript{out})}(r_{ij}).
\end{align*}

After obtaining the above weight as well as $\hat{\mathbf{x}}$ by rescaling the irreducible representations $\mathbf{x}$ with a norm gate followed by self-interaction, a tensor product filter is applied to produce the node pair irreducible representations $\mathbf{f}_{ij}$ as 
\begin{equation}
    \mathbf{f}_{ij}^{\ell} = (F^{(\ell\textsubscript{in}_i, \ell\textsubscript{in}_j, \ell\textsubscript{out})}(r_{ij}) \quad \hat{\mathbf{x}}_i^{\ell\textsubscript{in}_i} \otimes \hat{\mathbf{x}}^{\ell\textsubscript{in}_j}_j)^{\ell\textsubscript{out}}.
\end{equation}

\paragraph{Diagonal Pair.} Unlike non-diagonal pair, the diagonal pair fuses irreducible representations of the same node as illustrated in Figure~\ref{fig:message-passing}.C. 
First, two norm gates with self-interaction are applied to the irreducible representation $\mathbf{x}_i$ in parallel to produce two irreducible representations $\mathbf{xl}_i$ and $\mathbf{xr}_i$. Then, the pair representation $\mathbf{f}_{ii}^{\ell\textsubscript{out}}$ is calculated by a tensor product with learnable parameters $\mathbf{W}$ for each path $(\ell\textsubscript{in}_l, \ell\textsubscript{in}_r, \ell\textsubscript{out})$, which is defined as
\begin{equation}
    \mathbf{f}_{ii}^{\ell\textsubscript{out}} = (\mathbf{W}^{(\ell\textsubscript{in}_l, \ell\textsubscript{in}_r, \ell\textsubscript{out})} \quad 
    \mathbf{xl}_i^{\ell\textsubscript{in}_l} \otimes \mathbf{xr}_i^{\ell\textsubscript{in}_r})^{\ell\textsubscript{out}}.
\end{equation}

After that, a residual connection is added between the obtained diagonal pair representations and the input irreducible representations: 
\begin{equation}
    \mathbf{f}_{ii}^{\ell} = \mathbf{f}_{ii}^{\ell} + \mathbf{x}_i^{\ell}. 
\end{equation}

Finally, a norm gate followed by a self-interaction layer updates the pair features $\mathbf{f}_{ii}^{\ell}$ and outputs the final irreducible representations for diagonal pairs.

\begin{figure}[t]
    \vskip 0.2in
    \centering
    \includegraphics[width=0.45\textwidth]{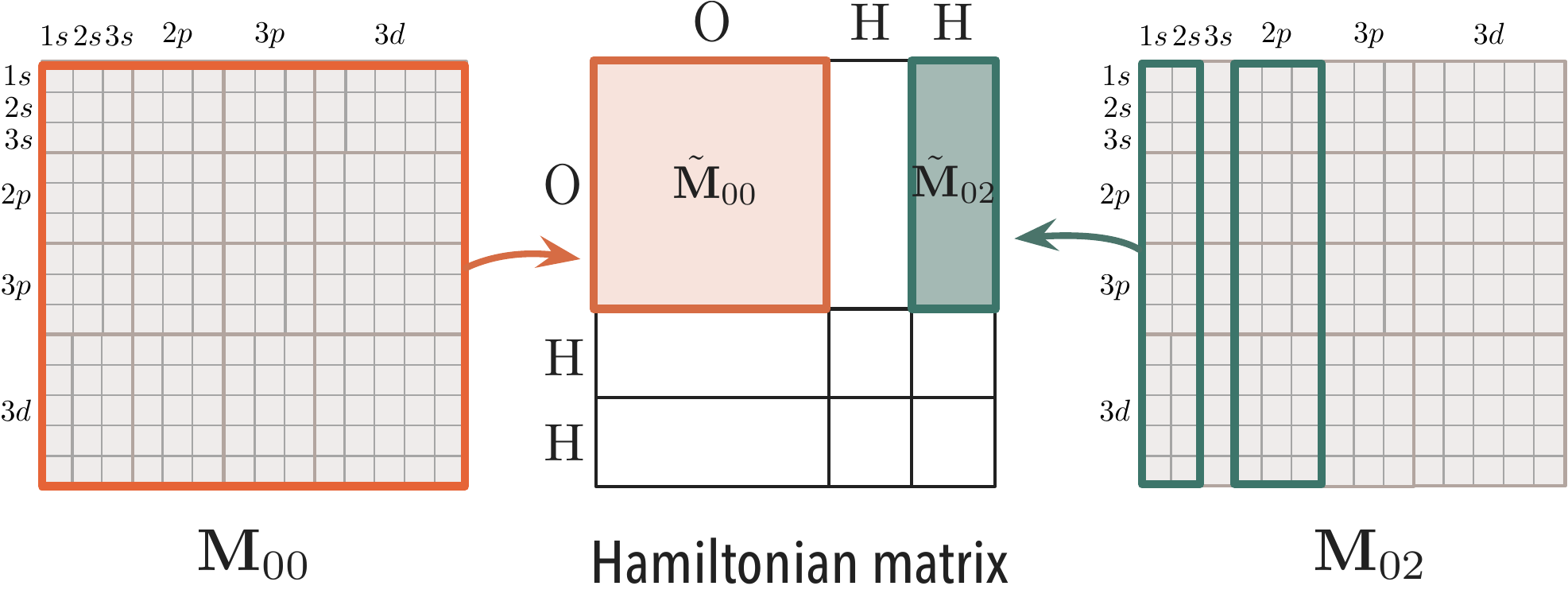}
    \caption{Construction of tensor matrix. The expansion module of our model outputs the intermediate block $\mathbf{M}$ with full orbitals for each node pair. Then, pair block $\tilde{\mathbf{M}}$ is selected from $\mathbf{M}$ according to orbitals of corresponding atoms. Finally, the Hamiltonian matrix of a molecule is built by combining $\tilde{\mathbf{M}}$ of all node pairs.}
    \vskip -0.2in
    \label{fig:build_matrix}
\end{figure}

\paragraph{Construction of Tensor Matrix.} After collecting irreducible representations for diagonal and non-diagonal pairs, the next step is to build the final Hamiltonian matrix. 
As illustrated in Figure \ref{fig:build_matrix}, each entry in the  matrix denotes the interaction between two orbitals. Specifically, pair block $\tilde{\mathbf{M}}_{ij}$ contains all the interactions between atoms $i$ and $j$. 
Since atoms have different numbers of orbitals, pair blocks $\tilde{\mathbf{M}}_{ij}$'s are in different shapes that should be determined when constructing the Hamiltonian matrix. 
In our framework, we introduce an intermediate block $\mathbf{M}_{ij}$ with full orbitals and then extract $\tilde{\mathbf{M}}_{ij}$ from that based on corresponding atom orbitals. For example, there are four atoms H, C, N, and O, in the MD17 dataset. In this case, full orbitals contain 1s, 2s, 3s, 2p, 3p, and 3d, and $\mathbf{M} \in \mathbb{R}^{14\times 14}$. When dealing with the hydrogen atom H, only 1s, 2s, 2p orbitals are selected to construct the Hamiltonian matrix. In this way, each node pair irreducible representation $\mathbf{f}_{ij}$ can be converted to an intermediate pair block $\mathbf{M}_{ij}$ with a predefined shape despite the atom type. This strategy is advantageous as it can be easily extended to different molecules.

\begin{figure}[t]
    \centering
    \vskip 0.2in
    \includegraphics[width=0.48\textwidth]{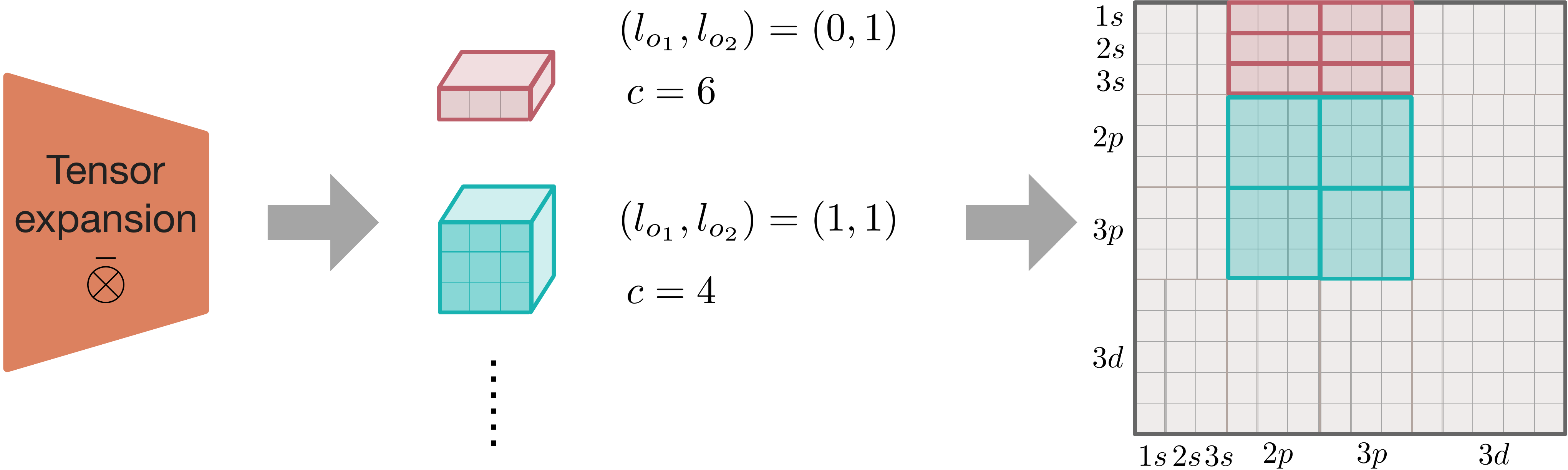}
    \caption{The relationship between the channel of equivariant matrix and the intermediate full orbital matrix. Here, c is the number of channel for equivariant matrices, and $(l_{o1}, l_{o2})$ is the rotation order of the equivariant matrix. Then, these equivariant matrices composed of the intermediate full orbital matrix for each node pair.}
    \vskip -0.2in
    \label{fig:expansion2matrix}
\end{figure}

\begin{figure}[t]
    \centering
    \vskip 0.2in
    \includegraphics[width=0.48\textwidth]{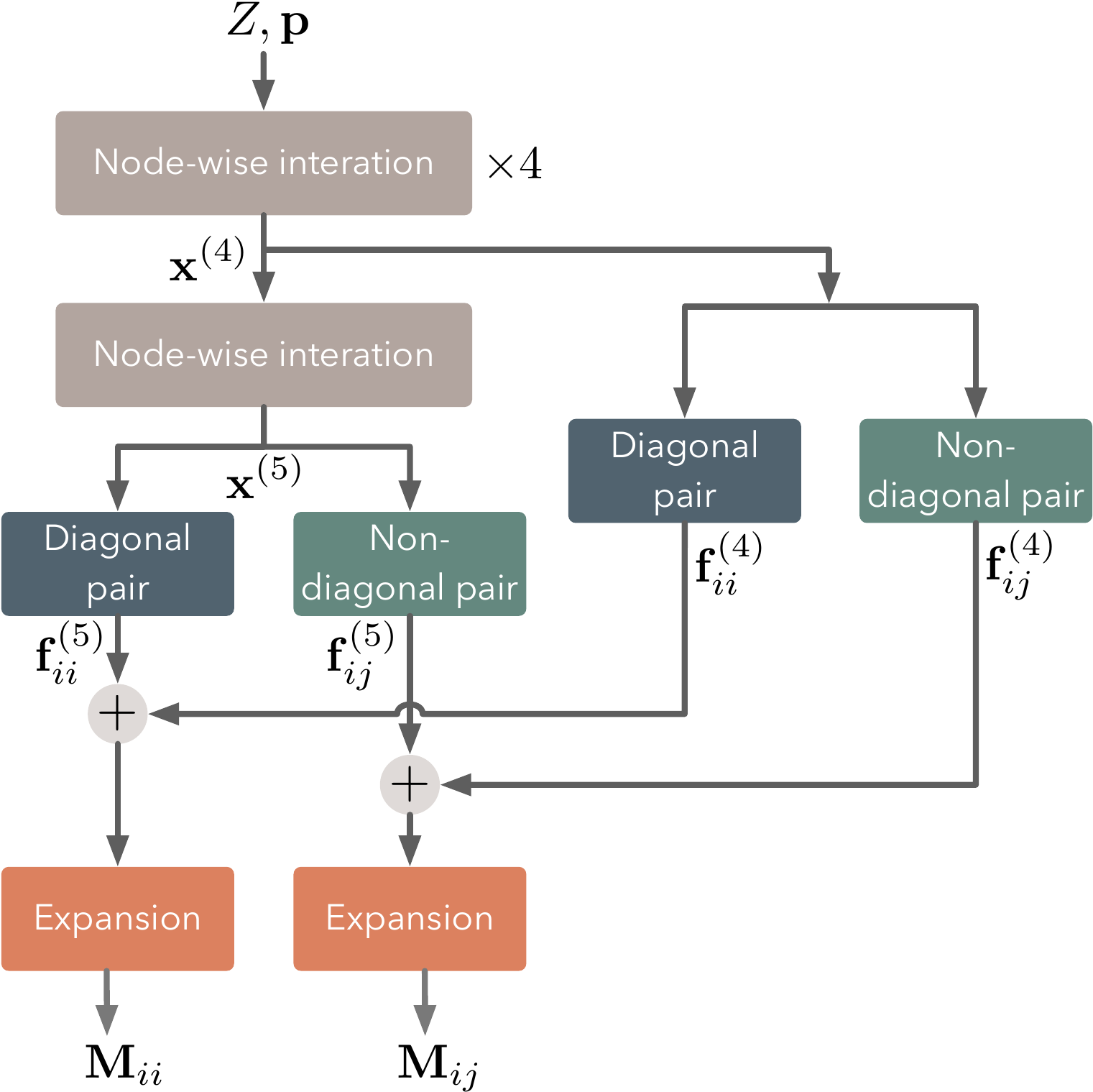}
    \label{fig:network}
        \caption{The whole architecture of the proposed QHNet. Inputs of the QHNet include atom types $Z$ and atom positions $\mathbf{p}$. Five layers of node-wise interaction are used to learn SE(3)-equivariant irreducible representations for atoms. Then, diagonal and non-diagonal atom pairs are passed to distinctive modules to create pairwise representations $\mathbf{f}_{ii}$ and $\mathbf{f}_{ij}$. Finally, the expansion module converts pairwise representations to the intermediate blocks $\mathbf{M}$ that are post-processed to build the quantum tensor.}
    \vskip -0.2in
\end{figure}

To construct the intermediate pair blocks with full orbitals using pair irreducible representations, we apply a tensor expansion operation with the filter operation as shown in Figure~\ref{fig:message-passing}.E. The tensor expansion is defined as
\begin{equation}
    (\bar{\otimes} w^{\ell_3})^{(\ell_1, \ell_2)}_{(m_1, m_2)} = 
            \sum_{m_3=-\ell_3}^{\ell_3}  C_{(\ell_3, m_3)}^{(\ell_1, m_1), (\ell_2, m_2)} w_{m_3}^{\ell_3},
\end{equation}
where $C$ is the CG matrix and $\bar{\otimes}$ denotes tensor expansion which is the inverse operation of tensor product. Since CG matrix can be transformed to a unitary matrix, when $w^{\ell_3} = (u^{\ell_1} \otimes v^{\ell_2})^{\ell_3}$, we deduce that $ u^{\ell_1} \otimes v^{\ell_2} = \sum_{\ell_3}\bar{\otimes} w^{\ell_3}$ with $|\ell_1 - \ell_2 | \leq l_3 \leq \ell_1 + \ell_2$.
Then, the filter takes atom types as input to produce a weight for each path $(\ell\textsubscript{o1}, \ell\textsubscript{o2}, \ell\textsubscript{in})$: 
\begin{align}
    F_{ij}^{(\ell\textsubscript{o1}, \ell\textsubscript{o2}, \ell\textsubscript{in})} &= f(Z_i, Z_j) \nonumber \\
    F_{ii}^{(\ell\textsubscript{o1}, \ell\textsubscript{o2}, \ell\textsubscript{in})} &= f(Z_i),
\end{align}
where $Z$ denotes the embedding of atom types. 
Finally, the intermediate blocks $\mathbf{M}$ are built by the filter and node pair irreducible representations as
\begin{align}
    M_{iic}^{(\ell\textsubscript{o1}, \ell\textsubscript{o2})} = \sum_{
    \ell\textsubscript{in}, c'} F_{iicc'}^{(\ell\textsubscript{o1}, \ell\textsubscript{o2}, \ell\textsubscript{in})} \bar{\otimes} \textbf{f}_{iic'}^{\ell\textsubscript{in}} \nonumber \\
    M_{ijc}^{(\ell\textsubscript{o1}, \ell\textsubscript{o2})} = \sum_{
    \ell\textsubscript{in}, c'} F_{ijcc'}^{(\ell\textsubscript{o1}, \ell\textsubscript{o2}, \ell\textsubscript{in})} \bar{\otimes} \textbf{f}_{ijc'}^{\ell\textsubscript{in}},
\end{align}
where $c'$ denotes the channel index in the input irreducible representations and $c$ denotes the channel index in $\mathbf{M}$. For example, there are nine channels with $(\ell\textsubscript{o1}, \ell\textsubscript{o2}) = (0, 0)$, and four channel with $(\ell\textsubscript{o1}, \ell\textsubscript{o2}) = (1, 1)$.
Note that a bias term is learned for node pair representation with $\ell\textsubscript{in} = 0$. We provide a demonstration showing the mapping from the output of expansion module to the full orbital matrices in Figure \ref{fig:expansion2matrix}.

In the tensor expansion module of PhiSNet~\cite{schutt2019unifying}, a record has to be maintained to map the relationship of atom-orbital pair interactions so that a unique channel index in irreducible representations can be selected for each interaction. In contrast to determining the channel dimension by the number of atom-orbital pair interactions in PhiSNet, our model does not need to keep the record because our channel dimension is fixed as a hyper-parameter. As a result, our model can effectively prevent the exponential growth of channel dimension when more atom types are involved.

\begin{table}[t]
\centering
\caption{Comparison of the total number and maximum sequential number of tensor products in PhiSNet and QHNet.}
\label{tab:tp-comparison}
\begin{tabular}{ccc}
    \toprule[1pt]
    Methods & \# of TP & \# of sequential TP \\ 
    \midrule
    PhiSNet &  121 & 76 \\
    QHNet    &  9 & 6 \\
    \bottomrule[1pt]
\end{tabular}
\end{table}

\begin{table*}[!h]

    \caption{Overall performance comparison in validation and test sets. Both PhiSNet and our QHNet are trained using a learning rate scheduler with a linearly decreasing learning rate to ensure the convergence with 1,000 warm-up steps and 200,000 total steps with double-precision floating-point. The unit for Hamiltonian $\mathbf{H}$ and eigen energies  $\boldsymbol{\epsilon}$ is Hartree, denoted by $E_h$. Training `Time' is in days. Training loss is described in Appendix~\ref{sub:trainingloss}.}
    \centering
\scalebox{0.92}{
\begin{tabular}{llccccccc}
  \toprule[1pt]
  \multirow{2}{*}{Dataset} & \multirow{2}{*}{Method} & \multirow{2}{*}{Time} & 
  \multicolumn{3}{c}{Validation} & \multicolumn{3}{c}{Test} \\ 
  &&&$\mathbf{H}$ $[10^{-6} E_h] \downarrow$  &  $\boldsymbol{\epsilon}$ $[10^{-6} E_h] \downarrow$  & $\psi$ $[10^{-2}] \uparrow$ & $\mathbf{H}$ $[10^{-6} E_h] \downarrow$ &  $\boldsymbol{\epsilon}$ $[10^{-6} E_h] \downarrow$ & $\psi $ $[10^{-2}] \uparrow$  \\
  \midrule
  \multirow{2}{*}{Water}& PhiSNet    & 4.67d &  7.87  & 29.81 & 99.99 & 15.67 & -- & 99.94 \\
  & QHNet   & 1.27d & 10.49 & 31.62 & 99.99 & 10.79 & 33.76 & 99.99 \\
  \midrule
  \multirow{2}{*}{Ethanol}   & PhiSNet    & 7.45d & 19.60  & 101.10 & 99.91 & 20.09 & 102.04 & 99.81 \\
  & QHNet   & 3.73d & 20.45 & 81.29 & 99.99 & 20.91 & 81.03 & 99.99 \\
  \midrule
  Malondial- & PhiSNet   & 8.75d &  21.89  & 104.76 & 99.96 & 21.31 & 100.60 & 99.89 \\
  dehyde& QHNet  & 3.33d &  22.07 & 86.04  & 99.89 & 21.52 & 82.12  & 99.92 \\
  \midrule
  \multirow{2}{*}{Uracil}    & PhiSNet   & 14.12d & 18.89  & 146.14 & 99.87 & 18.65 & 143.36 & 99.86 \\ 
  & QHNet  & 3.31d  & 20.33 & 113.48 & 99.88 & 20.12 & 113.44 & 99.89 \\
  \bottomrule[1pt]
\end{tabular}}
\vspace{5mm}
\label{tab:overall-performance}
\end{table*}

\begin{table*}[t]
    \caption{Comparing PhiSNet and QHNet in terms of \textbf{training time per iteration}  and \textbf{GPU memory consumption} on the MD17 datasets. Note that these are in double-precision floating-point.}
    \centering
    \begin{tabular}{lccccc}
    \toprule[1pt]
    Dataset   &  Batch Size & PhiSNet &  QHNet & Speedup &  Memory Efficiency\\
    \midrule
    Water           &  10 &  2.02s / 3,839M & 0.56s / 2,419M & 3.61$\times$ & 1.59$\times$\\
    Ethanol         &  5  &  3.22s / 8,673M & 0.84s / 3,917M & 3.83$\times$ & 2.21$\times$\\
    Malondialdehyde &  5  &  3.78s / 8,869M & 0.88s / 3,925M & 4.30$\times$ & 2.25$\times$\\
    Uracil          &  3  &  6.10s / 9,135M & 0.94s / 4,049M & 6.49$\times$ & 2.25$\times$\\
    \bottomrule[1pt]
    \end{tabular}
    \vspace{5mm}
    \label{tab:efficiency-comparison}
\end{table*}

\subsection{Model Architecture}
\label{sec:architecture}

We now describe our whole framework. 
The model takes atom types $Z \in\mathbb{N}^{N \times 1}$ and their positions $\mathbf{p} \in \mathbb{R}^{N \times 3}$ as inputs. Here $N$ denotes the number of atoms in a molecule.
As shown in Figure \ref{fig:network}, our network uses five node-wise interaction layers under the default setting to learn SE(3)-equivariant irreducible representations for atoms. 
Next, we take atom representations outputted by the last two layers to construct the diagonal and non-diagonal irreducible representations. Then, we combine the pairwise representations obtained after both layers with a sum operation and feed them into the expansion module.

The computation and time overheads of quantum tensor networks mainly come from message passing and tensor product~(TP) operations. The computational cost of TP is much larger than a linear layer because it involves multiplication with CG matrix for each path sequentially. 
The time complexity analysis for the forward procedure of our module is shown as following: linear $O(C^2)$, self-interaction $O(C^2L)$, linear layer $O(C^2L^2)$ in the norm gate, and tensor product $O(CL^6)$. Here, $C$ represents the number of channels, and $L$ denotes the maximum rotation orders. Notably, in our experiments with QHNet and PhiSNet, we set $C=128$ and $L=4$. Therefore, the tensor product operation is as the most time-consuming operation in the model.
In Table \ref{tab:tp-comparison}, we compare the total number of TP and the maximum number of sequential TP in our model and PhiSNet. The number of sequential TP is the number of passed TP for one variable from input to output. Since we use fewer TP operations, our model achieves higher efficiency  and consumes less GPU memory.

\section{Experiments}

\textbf{Dataset.} We conduct experiments to evaluate the performance of QHNet on MD17 datasets~\cite{schutt2019unifying}. MD17 consists of four datasets of small molecules, including water, ethanol, malondialdehyde, and uracil. Each dataset contains thousands of 3D molecular structures and their corresponding Hamiltonian matrices. The Hamiltonian matrices in MD17 were calculated by using the def2-SVP basis set~\cite{weigend2005balanced} with PBE~\cite{perdew1996generalized} density functionals. The statistics of MD17 dataset is shown in Table \ref{tab:statistic}.

\textbf{Software and Hardware.} Our experiments are implemented based on PyTorch 1.11.0~\cite{paszke2019pytorch}, PyTorch Geometric 2.1.0~\cite{Fey/Lenssen/2019}, and e3nn~\cite{e3nn_paper}. In our experiments, models are trained on a single 11GB Nvidia GeForce RTX 2080Ti GPU and Intel Xeon Gold 6248 CPU.

\subsection{Overall Performance Evaluation}
\label{sec:exp_performance}

\textbf{Setup.}
To assess the efficiency and efficacy of the proposed QHNet, we conduct experiments over four molecular systems collected in MD17 datasets, including water, ethanol, malondialdehyde, and uracil. 
Since the number of optimization steps plays an important role in model training on MD17, 
we set the total training steps to 200,000 for all experiments. 
We adopt the learning rate scheduler to speed up the convergence of model training.
Specifically, the scheduler increases the learning rate gradually during the first 1,000 warm-up steps. The initial learning rate is 0, and the maximum learning rate is $5e^{-4}$. Then, the scheduler reduces the learning rate linearly so that the learning rate reaches $1e^{-7}$ at the last step.
Our competing baseline PhiSNet has five Pairmix layers that incorporate the representations from neighbor nodes to update node representations in the equivariant atomic representation network. For a fair comparison, QHNet uses five node-wise interaction layers to aggregate the messages from neighbor nodes to update node irreducible representations. 

Note that the batch size differs for different molecules. Specifically,  for QHNet, the batch size is set to 10 for water, ethanol, and malondialdehyde, while it is set to 5 for uracil. On the other hand, for PhiSNet, the batch size is as follows: 10 for water, 5 for ethanol and malondialdehyde, and 3 for uracil. 

Model parameters are set in double-precision floating-point format when we run experiments for both the baseline and our QHNet. It is because the resolution of single-precision, $1e^{-6}$, is at a similar order of magnitude to the predicted error. Using double-precision can approximate at a higher resolution, and it provides more consistent and reliable evaluation with DFT algorithms where double-precision is required to ensure high accuracy. Note that we omit the task of predicting overlap matrix, since overlap matrix can be easily calculated without any errors in $10^{-3}s$ for a these molecules shown in Table \ref{tab:inference_time_comp}.

\textbf{Results.} We provide the overall performance results in Table~\ref{tab:overall-performance}. Three metrics were applied to evaluate the accuracy of the predicted Hamiltonian matrices, including mean absolute error~(MAE) of the Hamiltonian matrix elements $\mathbf{H}_{ij}$. 
Besides directly checking the prediction error of Hamiltonian matrices, it is also important to evaluate the error of the predicted orbital energy $\boldsymbol{\epsilon}$ and wavefunction $\psi$, which can be deduced from the Hamiltonian matrix $\mathbf{H}$ through Eqs. \eqref{eq:wavefunction} and~\eqref{eq:roothaan}. 
Therefore, we report the MAE of the predicted energies and cosine similarity of coefficients for occupied molecular orbitals.

As reported in Table~\ref{tab:overall-performance}, QHNet can achieve competitive MAE on $\mathbf{H}$ in terms of both validation and test accuracy, and obtain better results on orbital energies.
Specifically, on the water dataset, QHNet outperforms PhiSNet by a significant margin of $4.88 \times 10^{-5} E_h$ .
Although its obtained MAEs of predicted Hamiltonian matrices $\mathbf{H}$ are similar to the baseline, QHNet can better predict molecular properties from the predicted $\mathbf{H}$. For orbital energies, QHNet consistently outperforms other models on ethanol, malondialdehyde, and uracil datasets. We did not include the test MAE of $\boldsymbol{\epsilon}$ by PhiSNet on the water dataset in Table \ref{tab:overall-performance} as it is ten times worse than that by QHNet (33.76). 
All these experimental results demonstrate the efficacy and better generalizability of QHNet.


\subsection{Time and Memory Efficiency}
\label{sec:exp_efficiency}

We further compare the running time and GPU memory consumption of PhiSNet and QHNet. For each dataset, we set the same batch size for both models, and show the comparison in Table~\ref{tab:efficiency-comparison}.
The result indicates that the proposed QHNet is much more efficient than PhiSNet. Specifically, QHNet runs more than three times faster during training, requiring less than half of the GPU memory. As explained in Sec~\ref{sec:architecture} and shown in Table~\ref{tab:tp-comparison}, the high efficiency of our QHNet is because the total number and sequential number of tensor products are reduced to less than 14\% of that in PhiSNet. 

\subsection{Performance on Mixed Dataset}
Experiments in Sec~\ref{sec:exp_performance} and Sec~\ref{sec:exp_efficiency} focus on training and testing models on the same molecular system. Thus, the predicted Hamiltonian matrices have the same shape as training Hamiltonian matrices.
In this subsection, we mix previously mentioned four datasets together while keeping the original split of training, validation, and testing sets. In this case, the mixed dataset contains four kinds of molecules, and QHNet is trained to predict the Hamiltonian matrices for multiple molecules rather than one.
As described in Sec~\ref{sec:buil_block}, we adopt the design of full orbital matrices with a fixed shape in the expansion module. Hence, QHNet can be easily extended to this mixed dataset and still achieve comparable performances, as demonstrated in Table~\ref{tab:performace-on-mixed-dataset}. 
Note that such a mixed task is complex for PhiSNet with potential troubles for reasons we explain at the end of Sec~\ref{sec:buil_block}. The flexible implementation of QHNet facilitates training a universal quantum tensor prediction network to further advance deep learning for quantum chemistry and condensed matter physics.

\begin{table}[t!]
    \centering
    \caption{Performance of QHNet trained on the mixed dataset. Note that this dataset includes only 500 water examples in the training set while there are 25,000 examples each for ethanol, malondialdehyde and uracil shown in Table \ref{tab:statistic} in Appendix~\ref{sub:statistics}.}
    \scalebox{0.85}{
    \begin{tabular}{ccccc}
        \toprule[1pt]
        Dataset & $\mathbf{H}$ $[10^{-6} E_h] \downarrow$ &  $\boldsymbol{\epsilon}$ $[10^{-6} E_h] \downarrow$ & $\psi$ $[10^{-2}] \uparrow$\\
        \midrule
        Water   &  190.03 & 304.48 & 99.99\\
        Ethanol &  51.68  & 130.68 & 99.97\\
        Malondialdehyde &  36.79 & 101.14 & 99.89\\
        Uracil & 34.77 & 124.86 & 99.80\\
        Mixed Dataset & 83.12 & 173.86 & 99.92\\
        \bottomrule[1pt]
    \end{tabular}}   
    \label{tab:performace-on-mixed-dataset}
\end{table}


\section{Conclusion}
In this work, we presented a SE(3)-equivariant network---\textbf{QHNet}---to predict the Hamiltonian matrix with high accuracy and efficiency. QHNet is built carefully to maintain the underlying symmetry while eliminating 92\% of tensor product operations to make it super streamlined compared to existing methods. We evaluated QHNet using the MD17 datasets to show that it can accelerate the training by more than three times while using less than 50\% of the GPU memory. Additionally, our experimental results indicate that QHNet achieves competitive MAE on the Hamiltonian matrix and can derive more accurate molecular properties. Furthermore, by outputting full orbital matrices with a fixed shape and applying post-processing, QHNet is highly versatile and can be extended to datasets mixed with a variety of molecules. Therefore, we believe QHNet has great potential to efficiently predict accurate quantum tensors such as  Hamiltonian matrix in a wide range of molecular systems.

\section*{Acknowledgments}
This work was supported in part by National Science Foundation grants CCF-1553281, IIS-1812641, OAC-1835690, DMR-2119103, DMR-2103842, IIS-1908220, IIS-2006861, and IIS-2212419, and National Institutes of Health grant U01AG070112. Acknowledgment is made to the donors of the American Chemical Society Petroleum Research Fund for partial support of this research. 

\bibliography{dive,example_paper}
\bibliographystyle{icml2023}

\newpage
\appendix
\onecolumn
\section{Appendix.}
\subsection{Training Loss for QHNet}
\label{sub:trainingloss}
For training, we follow the implementation of PhiSNet \cite{unke2021se}, and use Mean Absolute Error~(MAE) and Root Mean Squared Error~(RMSE) as loss function to train the QHNet for predicting the Hamiltonian matrices.

\begin{equation}
    \mathcal{L}(\mathbf{H}, \mathbf{H}^{GT}) =  \left( \sqrt{\frac{1}{N^2} \sum_{i_1, i_2} (\mathbf{H}_{i_1, i_2} - \mathbf{H}^{GT}_{i_1, i_2})^2} + \frac{1}{N^2} \sum_{i_1, i_2} |\mathbf{H}_{i_1, i_2} - \mathbf{H}^{GT}_{i_1, i_2}| \right),
\end{equation}

where $\mathbf{H}$ is the predicted Hamiltonian matrix, $\mathbf{H}^{GT}$ is the ground-truth Hamiltonian matrix, and the size of  $\mathbf{\mathbf{H}}$  and $\mathbf{H}^{GT}$ is ${N}\times{N}$.

\subsection{Statistics of the dataset} \label{sub:statistics}

We here provide the statistics of the MD17 datasets in Table \ref{tab:statistic}.
Note that the mixed dataset follows the same dataset split.
\begin{table}[t]
    \caption{The statistics of MD17 dataset \cite{schutt2019unifying}.}
    \centering
    \scalebox{0.92}{
    \begin{tabular}{l|cccccccc}
    \toprule
        Dataset &  \# of structures & Train & Val & Test & \# of atoms & \# of orbitals & \# of occupied molecular orbitals\\
        \midrule
        Water           & 4,900     & 500       & 500 & 3,900   & 3     & 24    & 5 \\
        Ethanol         & 30,000    & 25,000    & 500 & 4,500   & 9     & 72    & 10 \\
        Malondialdehyde & 26,978    & 25,000    & 500 & 1,478   & 9     & 90    & 19 \\
        Uracil          & 30,000    & 25,000    & 500 & 4,500   & 12    & 132   & 26 \\
    \bottomrule
    \end{tabular}}
    \label{tab:statistic}
\end{table}

\subsection{Accelerating the DFT algorithm.}
To study the acceleration by using QHNet, we compare the inference time of QHNet with the time consumption of DFT algorithm. Here we use PySCF to conduct DFT algorithm with PBE correlation functional and def2-SVP basis set. Moreover, we select DIIS as the SCF algorithm for the DFT calculation.
In Table \ref{tab:inference_time_comp}, QHNet exhibited an acceleration of approximately 1000x on water, and approximately 300x on both ethanol and malondialdehyde. Note that we conduct our experiments on 11GB Nvidia GeForce RTX 2080Ti GPU and Intel Xeon Gold 6248 CPU.

\begin{table}[t]
    \centering
    \caption{Time comparison for single example among DFT algorithm, calculating overlap matrix, PhiSNet inference and QHNet inference. Note that we use batch size 64 to conduct inference for both PhiSNet and QHNet.}
    \begin{tabular}{l|cccc}
    \toprule
    dataset             & DFT[s]     & overlap [$10^{-4}$s] & PhiSNet [$10^{-2}$s] & QHNet [$10^{-2}$s] \\
    \midrule
     water              &  11.38  & 4.87 & 1.26 & 1.09 \\
     ethanol            &  25.11  & 8.71 & 8.83 & 7.11\\
     malondialdehyde    &  40.63  & 7.27 & 9.15 & 7.92 \\
     \bottomrule
    \end{tabular}
    \label{tab:inference_time_comp}
\end{table}

\begin{wraptable}{R}{7cm}
    \vspace{-0.8cm}
    \centering
    \caption{The ratio of optimization steps when taking predicted Hamiltonian matrices as the initial status of the DFT algorithm to accelerate the optimization.}
    \begin{tabular}{l|ccc}
    \toprule
    dataset         &  QHNet & PhiSNet \\
    \midrule
    water           & 0.494 & 0.497 \\
    ethanol         & 0.554 & 0.562 \\
    malondialdehyde & 0.562 & 0.567  \\
    \bottomrule
    \end{tabular}
    \label{tab:dft_acceleration_ratio}
\end{wraptable}

Next, we focus onoptimization steps ratio when continuing to optimize the predicted Hamiltonian matrices using the DFT algorithm.  It shows the time ratio to get the converged Hamiltonian matrix with machine learning model acceleration and reflects the quality of the predicted Hamiltonian matrices, with higher-quality matrices requiring fewer optimization steps. Note that we use PySCF on 50 random selected geometries in each molecular dataset to conduct DFT algorithm with PBE correlation functional and def2-SVP basis set, and select DIIS as the SCF algorithm for the DFT calculation.

\subsection{Performance comparison}
To compare with PhiSNet, we trained our QHNet using the same settings as PhiSNet, employing the Reduce Learning Rate On Plateau (RLROP) scheduler. We initialized the scheduler with a learning rate of $5e-4$ and continued training until it reached $1e^{-6}$. In order to limit the training steps, we set the maximum number of steps to $1,000,000$. The results are presented in Table \ref{tab:preformance_comparsion}. QHNet successfully converged on the water dataset. However, for the other three datasets, QHNet did not converge, and we therefore report the final results. Furthermore, we implemented a linear schedule with warmup, specifying the warmup steps and total training steps as LSW, and (10,000, 1,000,000) denotes a warmup period of 10,000 steps and a total training duration of 1,000,000 steps. Additionally, the batch size is set to 10 for all the experiments. Note that the results of PhiSNet comes from its original paper. For the $\psi$, since it is reported as $1.00$ in the original PhiSNet paper, it can not compare without enough accurate digits. Therefore, we omit the number here.

\begin{table*}[t]
    \caption{Performance of QHNet with various training steps compared to the reported results in PhiSNet. In these experiments, QHNet are in single-precision floating-point following the setting in original PhiSNet experiments. * denotes the training is converged and EMA is not used in this experiment.}
    \centering
    \begin{tabular}{lllccc}
    \toprule
    Dataset   &  Method & Training statgies & $\mathbf{H}$ $[10^{-6} E_h] \downarrow$  &  $\boldsymbol{\epsilon}$ $[10^{-6} E_h] \downarrow$  & $\psi$ $[10^{-2}] \uparrow$  \\
    \midrule
     \multirow{2}{*}{Water}           
                    & QHNet*     &  RLROP    &  10.36 & 36.21 & 99.99       \\
                    \cmidrule{2-6}
                    & PhiSNet   &  RLROP    & 17.59 &   85.53 & -                       \\
    \midrule
    \multirow{3}{*}{Ethanol} &  
                    QHNet   & LSW (10,000, 1,000,000)   & 12.78 & 62.97 & 99.99    \\
                   &QHNet   & RLROP   & 13.12 & 	51.80 &	99.99    \\
                    \cmidrule{2-6}
                    & PhiSNet   &  RLROP       &  12.15 &  62.75 & -       \\
    \midrule
    \multirow{3}{*}{Malondialdehyde}   & QHNet  & LSW (10,000, 1,000,000)   &   11.97 & 55.57 & 99.94 \\
          & QHNet  & RLROP & 13.18 & 51.54 & 99.95\\
                    \cmidrule{2-6}
                    & PhiSNet   &  RLROP       &  12.32 & 73.50 & -  \\
    \midrule
    \multirow{2}{*}{Uracil} & QHNet  & LSW (10,000, 1,000,000)       & 9.96 & 66.75 & 99.95 \\
    \cmidrule{2-6}
    & PhiSNet &  RLROP & 10.73 & 84.03 & - \\
    \bottomrule
    \label{tab:preformance_comparsion}
    \end{tabular}
\end{table*}

\subsection{Ablation studies on the model architecture}
To study the various modules in QHNet, we conduct experiments to compare the QHNet without attentive scores or NormGate in node-wise interaction layers and report the MAE of the Hamiltonian matrix on the test set in Table \ref{tab:ablaion}.
\begin{table}[t]
    \centering
    \caption{Ablation study of QHNet.}
    \begin{tabular}{lccc}
    \toprule
        Dataset    & full model 	& wo attentive 	& wo NormGate \\
        \midrule
        water             &  10.79 & 10.65 	& 10.91   \\
        ethanol           & 20.91 	& - 	& 39.26   \\
        malondialdehyde   & 21.52 	& 30.89 	& 40.79 \\
        \bottomrule
        \label{tab:ablaion}
    \end{tabular}
\end{table}


\end{document}